\documentclass{article}
\usepackage{spconf,amsmath,graphicx}
\usepackage[utf8]{inputenc}
\usepackage[T1]{fontenc}
\usepackage[english]{babel}
\usepackage{textcomp}
\usepackage{amsthm}
\usepackage{amsmath}
\usepackage{amssymb}
\usepackage{times}
\usepackage{soul}
\usepackage{epsfig}
\usepackage{amssymb}
\usepackage{multirow}
\usepackage{url,subfigure,graphicx,color,xcolor,booktabs,colortbl,threeparttable}
\usepackage{algorithm,algorithmic}
\usepackage{pifont}
\usepackage{enumitem}
\usepackage{array}
\usepackage{bm}
\usepackage{url}
\usepackage{ifpdf}
\usepackage{ifxetex}
\usepackage{caption}
\usepackage{cite}
\usepackage{fancyhdr}

\title{Exploration of visual prompt in Grounded pre-trained open-set detection}

\name{Qibo Chen, 
Weizhong Jin,
Shuchang Li,
Mengdi Liu,
Li Yu,
Jian Jiang,
Xiaozheng Wang}
\address{China Mobile(Zhejiang) Research $\&$ Innovation Institute}

\begin{document}

\fancypagestyle{copyright}{\fancyhf{}\renewcommand{\headrulewidth}{0pt}\fancyfoot[L]{\scriptsize \copyright Copyright 2024 IEEE. Published in ICASSP 2024 - 2024 IEEE International Conference on Acoustics, Speech and Signal Processing (ICASSP), scheduled for 14-19 April 2024 in Seoul, Korea. Personal use of this material is permitted. However, permission to reprint/republish this material for advertising or promotional purposes or for creating new collective works for resale or redistribution to servers or lists, or to reuse any copyrighted component of this work in other works, must be obtained from the IEEE. Contact: Manager, Copyrights and Permissions / IEEE Service Center / 445 Hoes Lane / P.O. Box 1331 / Piscataway, NJ 08855-1331, USA. Telephone: + Intl. 908-562-3966.}}

\thispagestyle{copyright}
\maketitle
\begin{abstract}
Text prompts are crucial for generalizing pre-trained open-set object detection models to new categories.
However, current methods for text prompts are limited as they require manual feedback when generalizing to new categories, which restricts their ability to model complex scenes, often leading to incorrect detection results. To address this limitation, we propose a novel visual prompt method that learns new category knowledge from a few labeled images, which generalizes the pre-trained detection model to the new category.
To allow visual prompts to represent new categories adequately, we propose a statistical-based prompt construction module that is not limited by predefined vocabulary lengths, thus allowing more vectors to be used when representing categories.
We further utilize the category dictionaries in the pre-training dataset to design task-specific similarity dictionaries, which make visual prompts more discriminative.
We evaluate the method on the ODinW\cite{li2022elevater} dataset and show that it outperforms existing prompt learning methods and performs more consistently in combinatorial inference.
\end{abstract}

\vspace{-1mm}
\section{Introduction}
\vspace{-1mm}
\label{sec:intro}

Closed-set object detection frameworks\cite{carion2020end,redmon2016you,ren2015faster,zhu2020deformable} are typically trained to predict a set of predefined categories, which cannot dynamically adapt to the changing requirements of downstream application scenes.
Recently, open-set detection models based on vision language pre-training have obtained significant attention and are considered an effective way to solve the closed-set detection problem, where detectors can adaptively generalize to new categories under the guidance of text prompts in this method\cite{gu2021open,kamath2021mdetr,li2022grounded,zhang2023simple,liu2023grounding,zhao2022omdet,feng2022promptdet}.
\begin{figure}[!th]
  \centering
  \centerline{\includegraphics[width=0.80\linewidth]{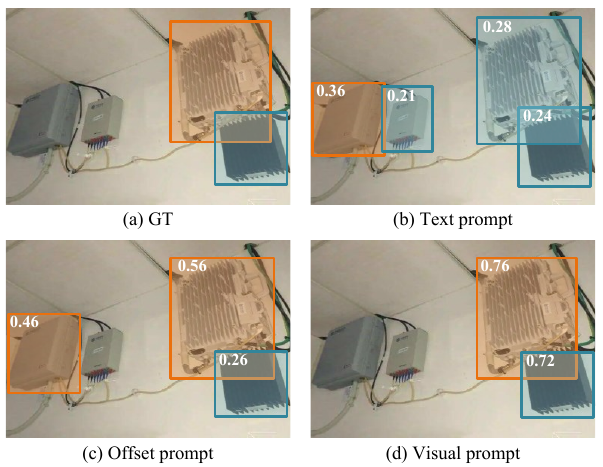}}
  \vspace{-2mm}
  \caption{\footnotesize {Visualization of different prompt combined inference. (a) Ground truth, orange is RRU(Radio Remote Unit), and cyan is Load. (b) Text prompts (category noun). (c) Learning offsets\cite{li2022grounded} of text prompts. (d) Our.}}
\vspace{-7mm}
\label{fig:intro_sp}
\end{figure}
Specifically, these methods use text prompt obtained from text encoder\cite{devlin2018bert,raffel2020exploring} to replace the classical parametric classification layer and train the detection model with large-scale object detection or language-image data\cite{shao2019objects365,kuznetsova2020open,plummer2015flickr30k} to align its image features and text prompt vectors.
Since natural language has intrinsic links, pre-trained models can use text descriptions of new categories to achieve generalization.
The most advanced models are GLIP\cite{li2022grounded} and Grounding DINO\cite{liu2023grounding}, which unify the object detection and phrase localization task formats to use the grounding dataset in pre-training. Both use text prompts (encoded text description) to design multi-layer cross-attention methods that realign the image features, further improving the detection model's understanding of text prompts.

\begin{figure*}[t]
	\begin{center}
		\includegraphics[width=0.8\linewidth]{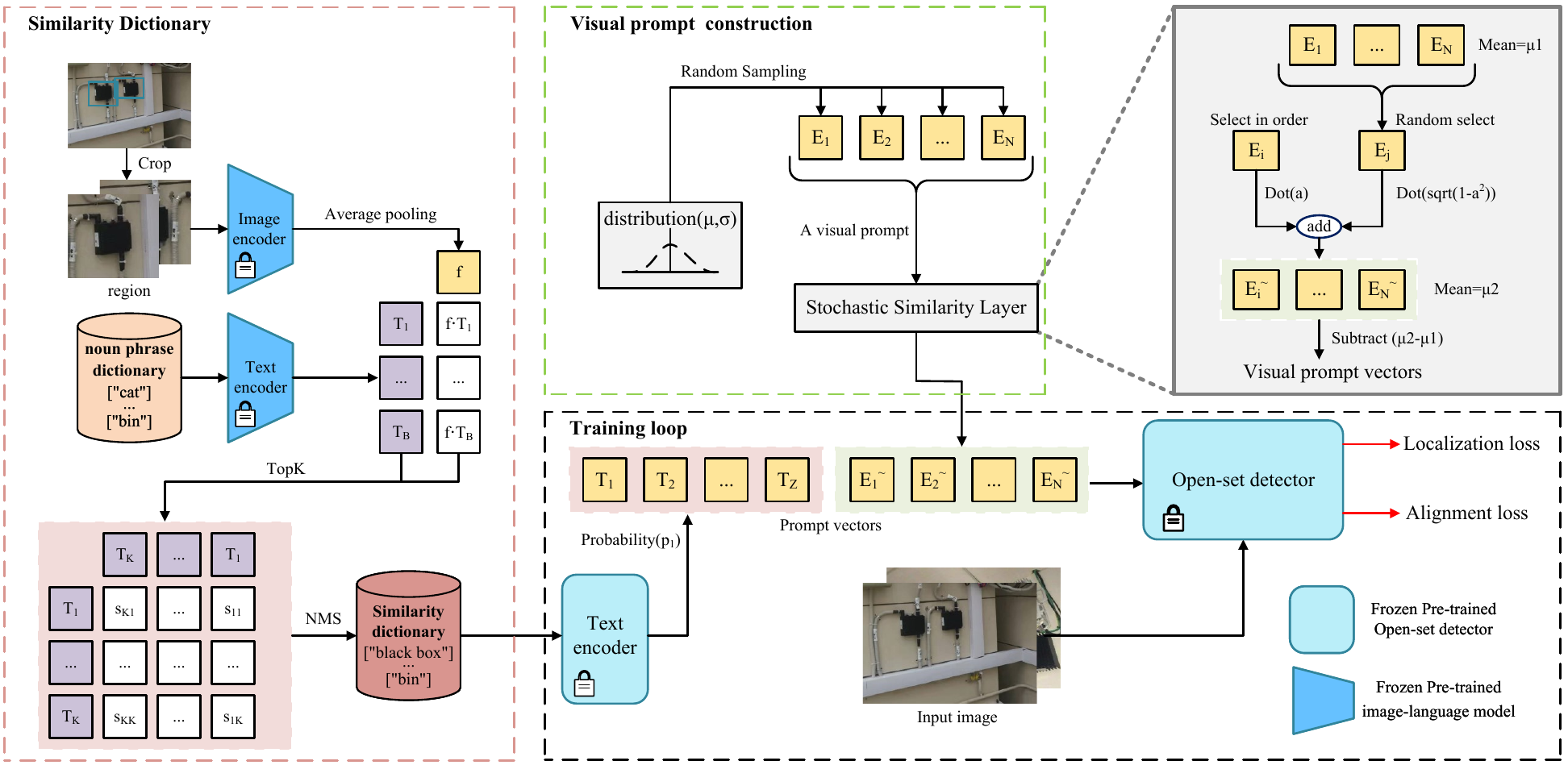}\\
	\end{center}
	\vspace{-0.25in}
	\caption{\footnotesize Illustration of our approach. During training, a visual prompt will concat with similar text prompts, as shown in the black dashed box, and then the loss is computed by the frozen detector. The green dashed box section is only used for visual prompt initialization. The red dashed box shows the construction process of the similarity dictionary, which selects top K texts based on similarity and saves them after removing duplicates via NMS.}
	\vspace{-0.2in}
	\label{fig2}
\end{figure*}

However, when inferencing with an image-language pre-training detector, we find serious limitations in using text prompts. The limitations can be categorized into the following three points: 
\textbf{1. Description difficulty}, many objects such as the specific device in Fig.~\ref{fig:intro_sp} (a) are difficult to be fully expressed in language. 
\textbf{2. Language ambiguity}, the same language description can refer to different objects\cite{yao2022detclip}, which can easily lead to incorrect detection in scenes with complex backgrounds, as shown in Fig.~\ref{fig:intro_sp} (b).
\textbf{3. Optimization uncertainty}, users and models have a gap in understanding the language, so manually modified text descriptions may lead to uncertain detection results.

Prompt learning may be a practical solution that maintains the pre-training capability of the model and optimizes the results stably with a few annotations.
This fine-tuning strategy originated in NLP\cite{li2021prefix,liu2021gpt} by using soft text prompts, which are learn-able vectors, as a task-specific context description and optimizing the prompt through annotation to adjust the model output.
In pre-trained image language classification, methods\cite{cho2021unifying,zhou2022learning,tsimpoukelli2021multimodal} also employ learning context by combining text prompt and learn-able vectors to improve the classification ability.
PromptDet\cite{feng2022promptdet} extends this idea to open-set detection by prompt learning the context representation better to align text prompts with the region image features. 
In contrast, GLIP\cite{li2022grounded} and OmDet\cite{zhao2022omdet} modify text prompts by learning the offset of text prompts to improve the detection accuracy of specific categories.
However, the above prompt learning methods in open-set detection still need humans to give text descriptions of the category and cannot avoid the problem of description difficulty.
Meanwhile, since the fine-tuning data is much smaller than the pre-training data, these methods lack sufficient negative samples when training the prompts, resulting in poor discriminative properties of the obtained prompts, which is manifested in the performance degradation of the prompts trained in different tasks when they are used together.
To illustrate, we followed \cite{li2022grounded} and trained the prompts separately, and the detection of Fig.\ref{fig:intro_sp} (c) is incorrect when inference together. 

After the above issues, we propose a novel visual prompt scheme free from the dependence on text description. 
Specifically, we first design the visual prompt construction module, which initializes visual prompts using statistical prior and stochastic similarity layer without manually setting text descriptions. 
Second, a similarity dictionary is constructed for downstream-specific tasks, which collects confusing text descriptions by pre-training the image-language model\cite{radford2021learning}. Adding similar text prompts in the visual prompt training reduces the ambiguity of visual prompts, and the stability of different-task visual prompt combined inference is effectively improved.
The visual prompts obtained by training in the above way can not only be combined but also be compatible with text prompts, so we only need to deploy the base model once to satisfy the downstream zero-shot demand and the high-precision demand in a specific scene, which significantly reduces the model deployment cost.
Our main contributions to this paper are:
\vspace{-2mm}
\begin{itemize}[leftmargin=*]
\item We propose a statistical-based visual prompt construction method that is not dependent on text descriptions and greatly extends the ability of the prompt to represent a new category.
\vspace{-1mm}
\item We propose a similarity dictionary strategy to make the learned image prompt more discriminative and significantly improve the stability of combined inference by adding confusing text prompts as negative samples during training.
\vspace{-1mm}
\item Our visual prompt method achieves state-of-the-art results on 13 public datasets compared to other prompt learning methods. In addition, ablation experiments show that our visual prompt runs well in combined inference.
\vspace{-2mm}
\end{itemize}

\vspace{-4mm}
\section{Methodology}
\label{sec:method}
\vspace{-2mm}
This paper develops an efficient visual prompt pipeline to remedy weaknesses in text prompts. 
The pipeline, illustrated in Fig.\ref{fig2}, uses visual prompt(consists of {\small$N$} vectors) to represent category and has three main components: visual prompt construction, similarity dictionary, and training loop.
For visual prompt construction(details in Sec.\ref{construction}), we first sample vectors independently from distribution as the initial visual prompt. We then use a stochastic similarity layer to correlate the within-class vectors while keeping between-class vectors uncorrelated. This eliminates reliance on manual text descriptions and allows the detector to distinguish the category to which the vectors belong early in optimization.
The similarity dictionary(details in Sec.\ref{dictionary}) is derived from noun phrases in the pre-training data by text-region similarity. During training, it generates confusing text prompts as negative examples to improve the discriminative representation of the visual prompts.
Finally, in the training loop, visual prompts are concatenated with negative prompts(lengths 0 to {\small$Z$}) at probability {\small$p_1$} to form the final prompts, which are fed into the frozen pre-trained open-set detector\cite{liu2023grounding} alongside the input image to produce detections. The visual prompt parts of the final prompts are jointly optimized through an alignment loss and localization loss.

\vspace{-4mm}
\subsection{Visual Prompt Construction}
\vspace{-1mm}
\label{construction}
As shown in Fig.\ref{fig2} green dashed box, the visual prompt {\small$F_v=\left\{E_i,\cdots,E_N\right\}$},{\small$E\in R^{1\times C}$} is used to represent single category, where {\small$C$} is determined by the pre-trained model and {\small$N$} is a customized value, so our visual prompt can improve the information capacity by increasing {\small$N$}.

\noindent {\textbf{Statistical Distribution Sampling.}}~The vectors of visual prompt were initialized by random sampling from Gaussian prior distributions.   
Specifically, we use the Natural Language Toolkit to extract noun phrases from the pre-trained labels and count the mean {\small$\mu$} and variance {\small$\sigma$} of the text prompts as priors by the detector's text encoder.

\noindent {\textbf{Stochastic Similarity.}}~Different visual prompts were processed separately by a stochastic similarity layer.
It first selects vector {\small$E_i$} in order from the visual prompt {\small$F_v$}, and then randomly selects {\small$E_j$} from the remaining vectors with probability {\small$p_2$}.
Then, the similarity is established by weighted fusion between vectors, which is defined as
\vspace{-2mm}
\begin{equation}\small
E_i^\sim={a E}_i+{\sqrt{(1-a^2)}E}_j,
\vspace{-2mm}
\label{eq1}
\end{equation}
where {\small$a\in[0,1]$} is the coefficient of independence, and when {\small$a$} is smaller, the vector is more similar to each other. 
Since we use a Gaussian distribution, Eq.\ref{eq1} ensures that the variance is constant, but to make the mean constant, we utilize mean({\small$\mu_1,\mu_2$}) before and after processing to modify, which is defined as
\vspace{-2mm}
\begin{equation}\small
F_{v}-=(\mu_2-\mu_1),
\vspace{-2mm}
\label{eq2}
\end{equation}

\vspace{-3mm}
\subsection{Similarity Dictionary}
\vspace{-1mm}
\label{dictionary}
To avoid the detector confusing objects similar to the target category, we filter the text descriptions with a similarity matrix from the noun phrase dictionary of the pre-training data to construct the similarity dictionary, shown in Fig.\ref{fig2}.
Thanks to image and text space alignment during pre-training, text prompts in the similarity dictionary can be regarded as object features similar to the target category.

\noindent {\textbf{Text-region Similarity.}}~ To compute potentially similar noun phrases, we enlarge the labeled region twice and crop it out, use CLIP\cite{radford2021learning} image encoder to get several image features {\small$f_i$}, and obtain query features {\small$f$} by average pooling. 
Then we use text encoder to get text features {\small$\left\{T_1,\cdots,T_B\right\}$}, construct text-region similarity matrix by dot product with {\small$f$}, and select the top K as candidate texts, defined as follows
\vspace{-2mm}
\begin{equation}\small
\left\{T_1,\cdots,T_K\right\}=topK(f^T(T_1,\cdots,T_B)),
\vspace{-2mm}
\label{eq3}
\end{equation}

\noindent {\textbf{Remove Duplication.}}~Candidate text prompts by self dot product to construct a {\small$K{\times}K$} similarity matrix {\small$M_T$}. By setting similarity threshold {\small$q$}, we use NMS(Non-Maximum Suppression)\cite{neubeck2006efficient} to remove text with high repetitions among the {\small$K$} text prompts, finally obtaining the task-specific similarity dictionary.

\vspace{-3mm}
\subsection{Training Settings}
\vspace{-1mm}
\label{train}
\noindent {\textbf{Category Score.}}~In our method the visual prompt corresponding to category {\small$c$} contain {\small$\left\{E_1^\sim,\cdots,E_N^\sim\right\}$}, which we dot product with {\small$i$}-th image features {\small$F_i$} to get category similarity matrix {\small$W_ic$}.
\vspace{-2mm}
\begin{align}\small
&W_ic={F_i(E_1^\sim,\cdots,E_N^\sim)}^T,\\
&S_{ic}=\begin{cases}
        Gumbel\text{-}softmax(W_{ic}) \quad \text{Train}\\
        Max(W_{ic}) \quad \quad \quad \quad \quad \quad \quad \text{Eval}
        \end{cases}
\vspace{-2mm}
\label{eq4}
\end{align}
In order for all vectors to be optimized by the alignment Loss, we use gumbel-softmax\cite{jang2016categorical} with stochasticity to obtain the category score {\small$S_ic$} during training.

\noindent {\textbf{Alignment Loss.}}~Following \cite{li2022grounded,liu2023grounding}, our method uses binary sigmoid loss to align visual prompt and labeled region.
Since we randomly add negative text prompts from the similarity dictionary during training, which may appear in the image, the loss function is adjusted as Eq.\ref{eq5} to avoid optimization conflicts.
\vspace{-2mm}
\begin{align}\small
\mathcal{L}_{als}=\begin{cases}
        \frac{1}{D_p+D_n}\sum_{c=1}^{D_p+D_n}{y_c\log{{\sigma(S}_c)}} \quad \text{Positive}\\
        \frac{1}{D_p}\sum_{c=1}^{D_p}{y_c\log{{\sigma(S}_c)}} \quad \quad \quad \quad \text{Negative}
        \end{cases}
\vspace{-2mm}
\label{eq5}
\end{align}
where {\small$D_p$} is the number of visual prompts, {\small$D_n$} is the number of negative text prompts, {\small$y_c$} is label, and {\small$\sigma$} is sigmoid function. 

In addition to alignment loss, our method still uses localization loss, including box L1 loss and GIOU\cite{rezatofighi2019generalized} loss.

\vspace{-2mm}
\section{Experiments}
\label{sec:exp}
\vspace{-2mm}
\subsection{Dataset and Metric}
\vspace{-2mm}
ODinW\cite{li2022elevater} is a benchmark consisting of a series of real-world detection tasks, usually used to test the performance of models under real-world conditions.
Following previous work\cite{li2022grounded,liu2023grounding}, we used 13 of these datasets in our experiments, including PascalVOC, AerialDrone, Aquarium, Rabbits, EgoHands, Mushrooms, Packages, Raccoon, Shellfish, Vehicles, Pistols, Pothole and Thermal, as well as evaluated the algorithms using the mAP(IoU=0.50:0.95) and mAP50(IoU=0.50) metrics.

\vspace{-4mm}
\subsection{Implementation Details}
\vspace{-2mm}
To demonstrate our visual prompt method, we conducted experiments on state-of-the-art grounded pre-trained open-set detector\cite{liu2023grounding} and trained on AdamW for 12 epochs with 16 batch sizes.
The number of vectors {\small$N$} was set to 20 (limited by the detector and set to 10 in PascalVOC) in visual prompt, the threshold {\small$q$} for the similarity dictionary NMS was set to 0.7, the probabilities {\small$p_1$} and {\small$p_2$} are set to 0.7 and 0.5 respectively, the independence coefficient {\small$a$} was set to 0.99, and the maximum length {\small$Z$} of negative text prompts was set to 20.
We found that training visual prompts required a larger base learning rate than the original paper, which was set to 0.1 in our experiments.
The experiments are performed on an NVIDIA V100 GPU, and all other hyper-parameters are followed with the settings of the corresponding detectors.

\begin{table*}
\scriptsize
\centering
\setlength\tabcolsep{3.0pt} 
\caption{{\footnotesize Comparison with other prompt methods on ODinW13. The superscript $\dag$ indicates that the prompt method is initialized with "object" instead of the corresponding class name. The evaluation metric is mAP.}}
\vspace{-3mm}
\begin{tabular}{l|cccccccccccccc}
\hline
\multicolumn{1}{c|}{Method} & PascalVOC & AerialDrone & Aquarium & Rabbits & EgoHands & Mushrooms & Packages & Raccoon & Shellfish & Vehicles & Pistols & Pothole & Thermal & Avg   \\ \hline
Text prompt                 & 55.4     & 9.2       & 18.2    & 66.5   & 60.7    & 69.3     & 57     & 60     & 34.4     & 57.2    & 67.6   & 26.8   & 69.6   & 50.1 \\
Context Prompt\cite{feng2022promptdet}                   & 71      & 21.6       & 38.7    & 70.2   & 67.4    & 85.5     & 71.3    & 70.1   & 51      & 67.3    & 69.8   & 43    & 80.5   & 62.1 \\
Context Prompt$^\dag$                &64.8           & 19.1       & 25.7    & 70.4   & 66     & 84.6     & 68.6    & 71.3   & 40.2     &50.5          & 69.3   & 43.6   & 75.3   &57.6       \\
Offset Prompt\cite{li2022grounded}                        & 70.2     & 29.2       & 44.8    & 74    & 68.9    & 87.5     & 69.7    & 72.5   & 55.9     & 66.9    & 70.6   & 46.2   & 79.1   & 64.3 \\
Offset Prompt$^\dag$                     & 69.8     & 27.8       &43.5          &73.1         &67.6          &87.5           &67.7          &72.7         &52.1           &63.6          &70.1         & 43.7        & 80.3        &63       \\
Our visual prompt                         & 71.7     & 34.2       & 53    & 75.8   & 73.4    & 88.1     & 75.6    & 74.3   & 58.7     & 68     & 73.6   & 52.3   & 81.5   & 67.7 \\ \hline
\end{tabular}
\vspace{-6mm}
\label{tab1}
\end{table*}

\begin{table}[!ht]
\scriptsize 
\centering
\caption{{\footnotesize The ablation study of statistical distribution, stochastic similarity layer, and similarity dictionary.}}
\vspace{-0.1in}
\setlength{\tabcolsep}{1mm}{
\begin{tabular}{l|ccc|cc}
\hline
method   & \begin{tabular}[c]{@{}c@{}}Statistical \\ Distribution\end{tabular} & \begin{tabular}[c]{@{}c@{}}Stochastic \\ Similarity\end{tabular} & \begin{tabular}[c]{@{}c@{}}Similarity \\ Dictionary\end{tabular} & mAP  & mAP50 \\ \hline
visual prompt(A)     & \ding{55}                                                                   & \ding{55}                                                                & \ding{55}                                                                & 65.5 & 80.2  \\
visual prompt(B) & \checkmark                                                                   & \ding{55}                                                                & \ding{55}                                                                & 67 & 82.1  \\
visual prompt(C) & \ding{55}                                                                   & \checkmark                                                                & \ding{55}                                                                & 65.6   & 80.7  \\
visual prompt(D) & \ding{55}                                                                   & \ding{55}                                                                & \checkmark                                                                & 66.1 & 81.5  \\
visual prompt(E) & \checkmark                                                                   & \checkmark                                                                & \ding{55}                                                                & 67.3 & 82.9  \\
visual prompt  & \checkmark                                                                   & \checkmark                                                                & \checkmark                                                                & 67.7 & 83.5  \\ \hline
\end{tabular}}
\vspace{-4mm}
\label{table2}
\end{table}

\begin{table}[!ht]
\scriptsize 
\centering
\caption{{\footnotesize Explore combinatorial inference with two separately trained prompts.
Category A is Packages, and Category B is Pothole.}}
\vspace{-0.1in}
\setlength{\tabcolsep}{2mm}{
\begin{tabular}{l|c|cl|clcl}
\hline
\multicolumn{1}{c|}{method} & Similarity Dictionary & \multicolumn{2}{c|}{Category} & \multicolumn{2}{c}{mAP}   & \multicolumn{2}{c}{mAP50} \\ \hline
                            & \ding{55}                    & \multicolumn{2}{c|}{A}     & \multicolumn{2}{c}{72.6}     & \multicolumn{2}{c}{79.4}     \\
                            & \ding{55}                    & \multicolumn{2}{c|}{B}     & \multicolumn{2}{c}{51.9}     & \multicolumn{2}{c}{80.8}     \\
visual prompt               & \ding{55}                    & \multicolumn{2}{c|}{A+B}   & \multicolumn{2}{c}{54.2}     & \multicolumn{2}{c}{71.6}     \\
                            & \checkmark                  & \multicolumn{2}{c|}{A}     & \multicolumn{2}{c}{75.6} & \multicolumn{2}{c}{81.2} \\
                            & \checkmark                 & \multicolumn{2}{c|}{B}     & \multicolumn{2}{c}{52.3} & \multicolumn{2}{c}{80.8} \\
                            & \checkmark                  & \multicolumn{2}{c|}{A+B}   & \multicolumn{2}{c}{60.1} & \multicolumn{2}{c}{76.7} \\ \hline
\end{tabular}}
\vspace{-4mm}
\label{table3}
\end{table}

\vspace{-4mm}
\subsection{Results}
\vspace{-2mm}
All our experiments were conducted based on Grounding-DINO-T\cite{liu2023grounding}, which uses Swin-T as backbone, and pre-trained on O365\cite{shao2019objects365}, GoldG\cite{kamath2021mdetr}, Cap4M\cite{li2022grounded}.

\noindent \textbf{Comparison with other prompt}.~Our visual prompt method is compared with other prompt methods recently proposed of open-set detection on the 13 data of ODinw, as shown in Tab.~\ref{tab1}.
Text prompt uses the category description, which is the baseline for the prompt method.
Context prompt, from PromptDet\cite{feng2022promptdet}, learns the context vector of the text prompt. 
Offset prompt, from GLIP\cite{li2022grounded} and OmDet\cite{zhao2022omdet}, learns the offset of text prompt.
We also verified the impact of inaccurate text descriptions for text-dependent context and offset prompts. 
Specifically, we replaced the category description with the neutral noun "object" which will have $\dag$ superscript in Tab.~\ref{tab1}.
As seen in Tab.~\ref{tab1}, our visual prompts performed best on all datasets and achieved an average result of 67.7 mAP, which outperforms 17.6, 5.6, and 3.4 mAP compared with text prompt, context prompt, and offset prompt.

It is worth noting that the performance of the context prompt and offset prompt drops by 4.5 and 1.3 mAP, respectively, when using inaccurate text descriptions. In contrast, our visual prompts successfully escape the limitation of text descriptions.

\noindent \textbf{Ablation Study}.~To evaluate the effectiveness of our design, five visual prompt variants were designed depending on the module being used, and the results are shown in Tab.~\ref{table2}.
Visual prompt(A) was used as the baseline and initialized using the " object " text prompt with N set to 20.

\begin{table}[!ht]
\scriptsize 
\centering
\caption{{\footnotesize The ablation of vector length N in the visual prompt in ODinW13.}}
\vspace{-0.1in}
\setlength{\tabcolsep}{3mm}{
\begin{tabular}{l|ccccccc}
\hline
\multirow{2}{*}{Metrics} & \multicolumn{7}{c}{N length}                   \\ \cline{2-8} 
                         & 1    & 2    & 4    & 8    & 12   & 16   & 20   \\ \hline
mAP                      & 65.1 & 65.6 & 66.8 & 67.0 & 67.0 & 67.7 & 67.7 \\
mAP50                    & 79.3 & 79.9 & 81.2 & 82.7 & 83.3 & 83.1 & 83.5 \\ \hline
\end{tabular}
}
\vspace{-4mm}
\label{table4}
\end{table}

In Tab.~\ref{table2}, by comparing the visual prompt (A), (B), (C), (E), it was found that the use of statistical distribution and stochastic similarity layer gained 1.9, 0.5 mAP50, respectively, and the joint use resulted in a more significant gain of 2.7 mAP50, which indicates that the prompts construction module(Sec.\ref{construction} ) can effectively replace textual descriptions. 
Comparing visual prompt (A) and (D), a similar dictionary gained 0.6 mAP and 1.3 mAP50, and we consider these gains to come from the knowledge of pre-training data, which is implicitly used in the similarity dictionary.

To further validate our similarity dictionary, we choose the Packages and Pothole datasets in ODinW to train different categories of visual prompts individually and use the trained visual prompts in combination, and the results are shown in Tab.~\ref{table3}.
When the similarity dictionary was not used, the average accuracy of separate inference for category A and category B visual prompt was 62.3 mAP, and the combined inference resulted in an accuracy drop of 8.1 mAP. In contrast, when the similarity dictionary was used, the combined inference resulted in an accuracy drop of only 3.9 mAP, which demonstrated that the introduction of similarity negative prompts facilitated the learning of visual prompts with discriminative representations during training.
Therefore, our visual prompt method is more stable when combining inference, which helps to re-use prompts learning from different detection tasks.

Finally, we conducted ablation experiments for the vectors' length in a single visual prompt.
As shown in Tab.~\ref{table4}, the number of vectors and performance are positively correlated, while visual prompts are not text-dependent and can be efficiently utilized to achieve gains with long vectors.

\vspace{-3mm}
\section{Conclusion}
\vspace{-2mm}
\label{sec:conclusion}
In this paper, we present a novel visual prompt method for grounded pre-trained open-set object detection. We summarise the limitations of text prompts and propose a statistical-based prompt construction module to eliminate reliance on textual descriptions. We propose a similarity dictionary to improve representational discriminability by adding negative samples. Comparison and ablation experiments on public datasets validate the effectiveness of our approach. More importantly, visual prompts can be stably combined for inference, allowing multiple prompts to be reused in different downstream tasks.
In the future, we will explore visual prompts that do not require fine-tuning to reduce model maintenance and human cost further.
\vfill\pagebreak

{\small
\bibliographystyle{IEEEbib}
\bibliography{refs}}

\end{document}